%% file: ijcai26.tex
\documentclass{article}
\usepackage{ijcai26}
\usepackage{multirow}
\usepackage{times}
\usepackage{soul}
\usepackage{url}
\usepackage[hidelinks]{hyperref}
\usepackage[utf8]{inputenc}
\usepackage[small]{caption}
\usepackage{graphicx}
\usepackage{amsmath}
\usepackage{amsthm}
\usepackage{booktabs}
\usepackage{algorithm}
\usepackage{algorithmic}
\usepackage{listings}
\usepackage{xcolor}
\usepackage{enumitem}
\usepackage[most]{tcolorbox}
\usepackage[switch]{lineno}
\usepackage{natbib}

\newcommand{\tgsrag}{\textbf{TGS-RAG}}

\title{Text-Graph Synergy: \\ A Bidirectional Verification and Completion Framework for RAG}

\author{
Jiarui Zhong
\and
Hong Cai Chen\thanks{Corresponding author.}\\
\affiliations
School of Automation, Southeast University, Nanjing 210096, China\\
\emails
220245143@seu.edu.cn,
chenhc@seu.edu.cn
}

\begin{document}

\maketitle

\begin{abstract}
    Retrieval-Augmented Generation (RAG) has become a core paradigm for enhancing factual grounding and multi-hop reasoning in Large Language Models (LLMs). Traditional text-based RAG often retrieves logically irrelevant pseudo-evidence, while graph-based RAG is frequently hindered by search-time pruning, which may discard potentially valid reasoning paths. Existing hybrid approaches primarily adopt simple evidence concatenation or unidirectional enhancement, which fails to address the fundamental "Information Island" problem caused by asymmetric reasoning flows between unstructured text and structured graphs. We propose \textbf{TGS-RAG}, a unified framework for \textbf{T}ext-\textbf{G}raph \textbf{S}ynergistic enhancement. TGS-RAG introduces a bidirectional mechanism: (i) a \textbf{Graph-to-Text} channel that employs a Global Voting strategy from visited graph nodes to re-rank and refine textual evidence, filtering out semantic noise; and (ii) a \textbf{Text-to-Graph} channel that utilizes the \textbf{Memory-based Orphan Entity Bridging} algorithm. This algorithm utilizes textual cues to proactively resurrect valid but previously pruned reasoning paths from the search history without additional database overhead. Experimental results on multiple multi-hop reasoning benchmarks demonstrate that TGS-RAG significantly outperforms state-of-the-art baselines, achieving a superior balance between retrieval precision and computational efficiency.
\end{abstract}

\begin{figure}[t]
    \centering
    \includegraphics[width=\linewidth]{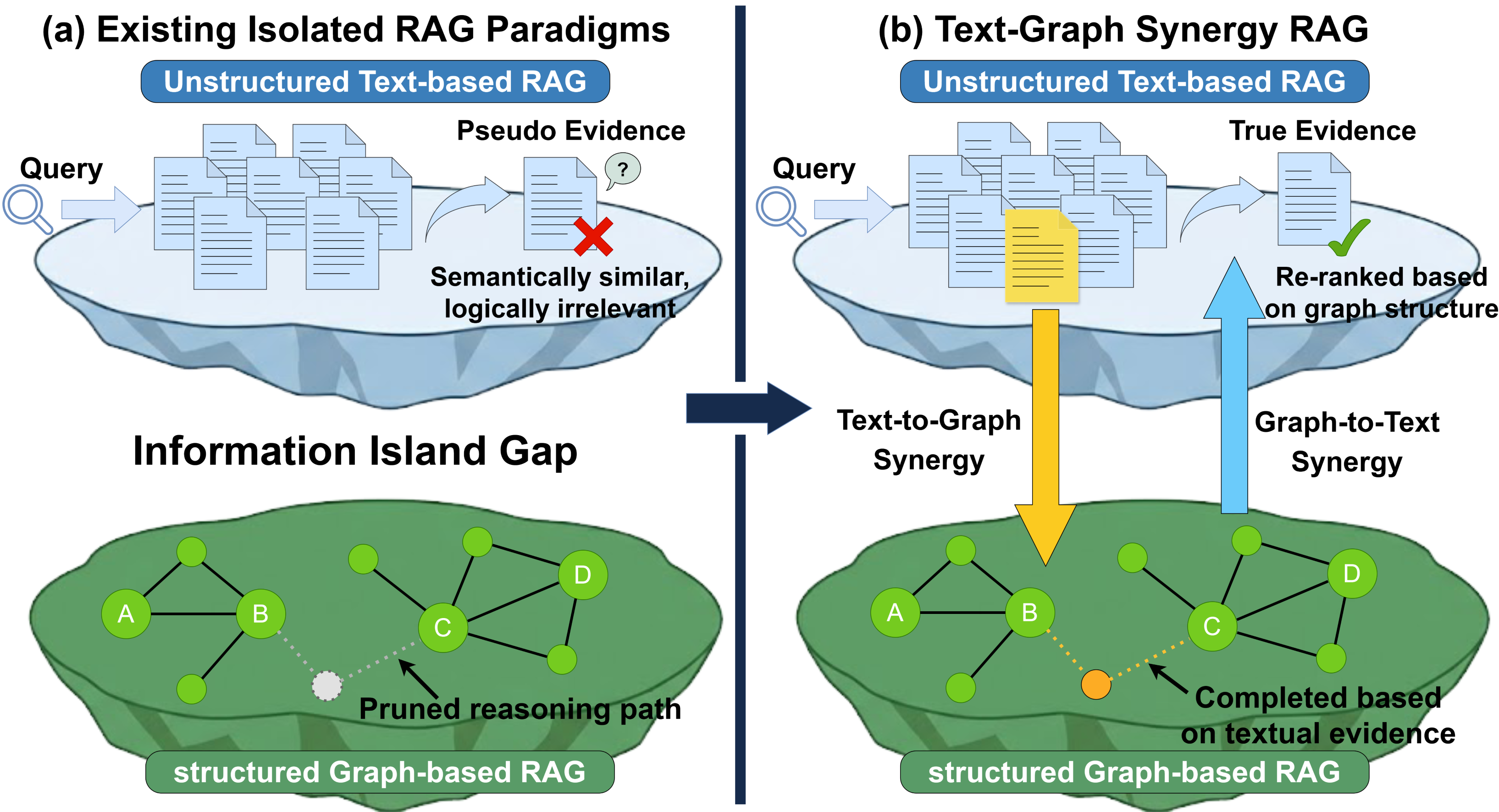}
    \caption{Comparison between isolated retrieval paradigms and the TGS-RAG framework. (a) Existing paradigms suffer from the "Information Island" gap: text-based methods often retrieve semantically similar but logically irrelevant \textit{pseudo-evidence}, while graph-based methods are hindered by \textit{broken reasoning paths} due to search-time pruning. (b) TGS-RAG bridges this gap through a bidirectional synergy mechanism, where graph structure guides the re-ranking of text to identify true evidence, and textual context facilitates the recovery of pruned but potentially valid reasoning paths.}
    \label{fig:motivation}
\end{figure}

\section{Introduction}

Large Language Models (LLMs) have achieved remarkable performance in natural language understanding and generation across a wide range of tasks. However, despite their impressive parametric capacity, LLMs remain fundamentally limited by their reliance on static internal knowledge and their tendency to produce hallucinated or logically inconsistent responses \citep{Ji_Survey_NLG2023}, particularly in knowledge-intensive and multi-hop reasoning scenarios. Retrieval-Augmented Generation (RAG) has therefore emerged as a central paradigm for grounding LLM outputs in external, verifiable evidence, enabling more factual, controllable, and interpretable generation \citep{NEURIPS2020_6b493230}.

Existing RAG systems predominantly follow two distinct paradigms. \textbf{Text-based RAG} retrieves relevant information from large unstructured corpora using dense vector similarity. While this paradigm offers high coverage, it is prone to retrieving pseudo-evidence \citep{Cossio2025}—textual chunks that are semantically similar to the query but logically irrelevant, and struggle with complex multi-hop reasoning. In contrast, \textbf{Graph-based RAG} leverages structured Knowledge Graphs (KGs) to provide high logical interpretability. However, these systems are often constrained by both structural sparsity in KGs and search-time pruning that may discard potentially useful reasoning paths.

Recent research attempts to combine Text-based RAG and Graph-based RAG to exploit their complementary strengths. However, these approaches often rely on simple evidence concatenation or pipeline-style augmentation, which treat textual and graphical evidence as independent sources, failing to achieve deep integration. Furthermore, unidirectional enhancement frameworks (e.g., KG-infused RAG) typically utilize one modality merely as an auxiliary signal for the other. Such designs fail to establish a closed-loop interaction, leaving the deeper potential of mutual verification and co-discovery largely unexplored.

At the core of this limitation lies the \textbf{Information Island problem}: textual evidence and graphical evidence are retrieved and processed in isolation, without a mechanism for mutual validation or collaborative reasoning. Specifically, current systems lack a principled framework in which (i) the structured logical constraints encoded in graphs can guide and refine text retrieval, while simultaneously (ii) the rich contextual cues present in unstructured text can validate, enrich, and recover pruned but potentially useful graph reasoning paths. In practice, this often means that the text and graph channels retrieve evidence that appears relevant on their own, but fails to mutually support a coherent reasoning chain.

To address these challenges, we propose \textbf{Text-Graph Synergy RAG (TGS-RAG)}\footnote{\url{https://github.com/EvannZhongg/TGS_RAG.git}}, a novel framework that enables deep, bidirectional integration between unstructured text and structured knowledge graphs during retrieval. TGS-RAG introduces a bidirectional enhancement mechanism: a \emph{Graph-to-Text} channel that utilizes structured logic to re-rank textual chunks, and a \emph{Text-to-Graph} channel that leverages contextual clues to validate and complete KG paths. Our main contributions are as follows:
\begin{itemize}
    \item \textbf{Bidirectional Synergy Framework:} We propose TGS-RAG, a unified framework that breaks the "information island" barrier by establishing a closed-loop feedback mechanism: graph structures guide the re-ranking of textual evidence (Graph-to-Text), while textual context validates and repairs graph reasoning paths (Text-to-Graph). This closed-loop design allows the two channels to move beyond simple coexistence and mutually verify the evidence retrieved by each other.
    
    \item \textbf{Memory-based Orphan Entity Bridging:} We introduce a novel algorithm that addresses the loss of potentially useful reasoning paths during beam-search pruning by treating pruned nodes as a \textit{deferred reasoning memory}. By leveraging textual cues to "resurrect" these orphan entities, we recover potentially valid reasoning paths discarded during initial search.
    
    \item \textbf{Cost-Effective Dual-Channel Reasoning:} We design a synergistic scoring mechanism that combines semantic similarity with structural voting. This approach effectively filters out pseudo-evidence and achieves an optimal Pareto frontier, delivering state-of-the-art accuracy with significantly lower token consumption than global graph indexing methods.
    
    \item \textbf{Empirical Superiority:} Extensive experiments on MuSiQue and HotpotQA demonstrate that TGS-RAG significantly outperforms existing text-based, graph-based, and hybrid baselines, particularly in scenarios requiring complex multi-hop reasoning across disjointed evidence.
\end{itemize}

\begin{figure*}[t]
    \centering
    \includegraphics[width=\textwidth]{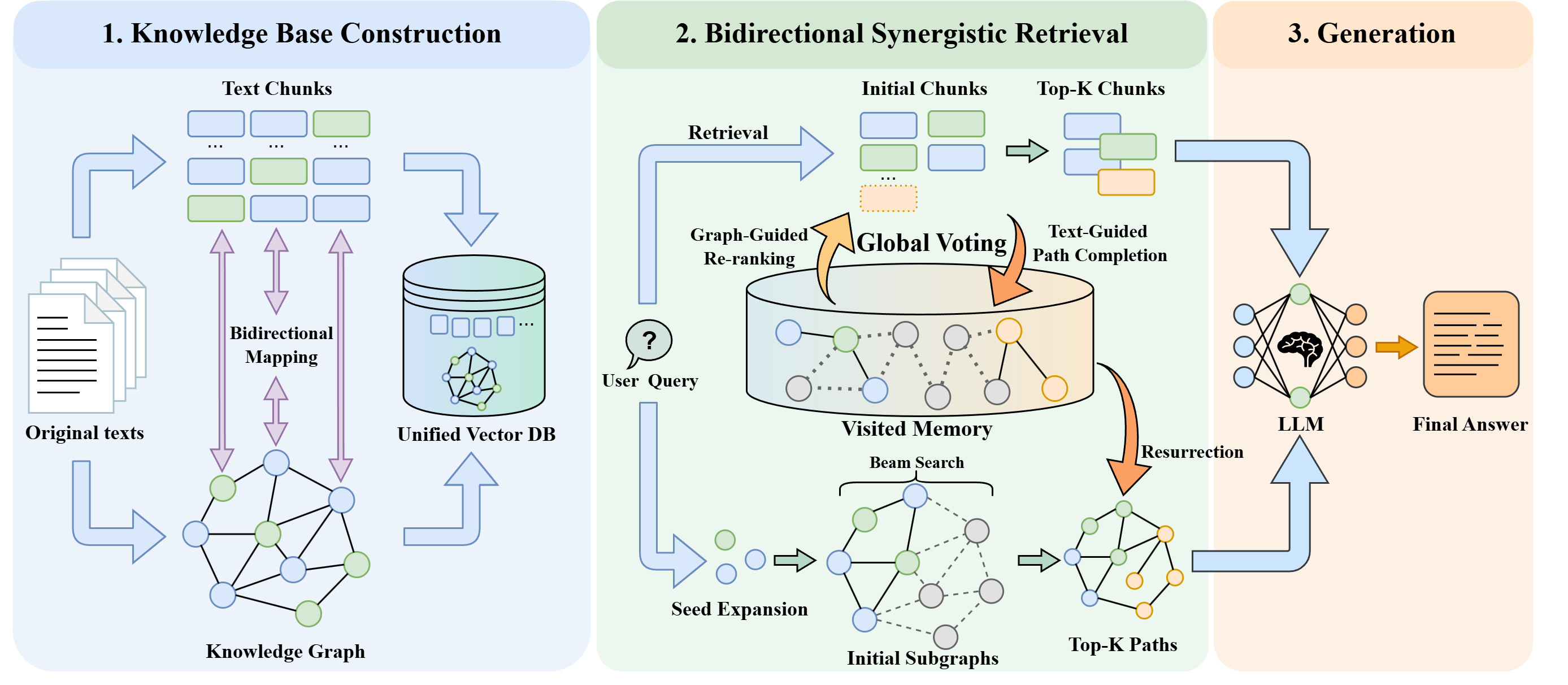}
    \caption{The overall architecture of TGS-RAG. The framework operates in three phases: (1) \textbf{Knowledge Base Construction}, where text chunks and entities are mapped in a unified vector DB; (2) \textbf{Bidirectional Synergistic Retrieval}, the core stage utilizing \textit{Semantic Beam Search} to accumulate a \textbf{Visited Memory} containing both selected and pruned nodes. This memory facilitates \textit{Graph-Guided Re-ranking} via \textbf{Global Voting} and enables \textit{Text-Guided Path Completion} by \textbf{resurrecting} pruned orphan entities from memory to bridge logical gaps; and (3) \textbf{Generation}, synthesizing the refined evidence.}
    \label{fig:framework}
\end{figure*}

\section{Related Work}

The evolution of RAG has transitioned from simple text matching to complex multi-modal and structural fusion. Our work builds upon three main research areas.

\subsection{Unstructured Retrieval for RAG}
Traditional RAG systems primarily rely on unstructured retrieval, where dense vector representations facilitate semantic matching between queries and documents. Dense Passage Retrieval (DPR) \cite{karpukhin-etal-2020-dense} pioneered the dual-encoder architecture for open-domain question answering, significantly improving recall over lexical methods. Subsequent models like ColBERT \cite{khattab2020colbert} introduced late interaction mechanisms to capture fine-grained token-level similarities. While these models excel at broad knowledge coverage, they often struggle with multi-hop queries where semantic similarity does not necessarily imply logical relevance, leading to the retrieval of misleading "pseudo-evidence."

\subsection{Structured Retrieval for RAG}
To address the logical limitations of text-based retrieval, recent studies have explored Knowledge Graphs (KGs) as structured knowledge sources. Methods in this category typically extract subgraphs or relational paths to provide deterministic evidence for LLMs. For instance, \textbf{G-Retriever} \cite{NEURIPS2024_efaf1c97} employs a GNN-based adapter to filter irrelevant nodes and edges, targeting efficient subgraph retrieval for textual graph understanding. Similarly, KG-based reasoning frameworks like ToG \cite{sun2024thinkongraph} utilize LLMs as agents to execute path-searching algorithms (e.g., beam search) over structured facts. Beyond path reasoning, recent graph-based indexing approaches leverage LLM-generated structures to capture global information. \textbf{GraphRAG} \cite{edge2024local} constructs hierarchical community summaries to answer global queries but suffers from high indexing costs. In contrast, \textbf{LightRAG} \cite{guo-etal-2025-lightrag} introduces a dual-level retrieval paradigm incorporating both graph structures and vector representations. However, these indexing-focused approaches primarily address retrieval efficiency and coverage, but are less equipped to handle scenarios where potentially useful reasoning paths are discarded during search.

\subsection{Text-Graph Integration Paradigms}
The integration of text and graphs has become a promising direction for robust RAG systems. Early approaches typically employed unidirectional enhancement strategies. For instance, KG-Infused RAG utilizes graph entities to expand queries for text retrieval, while frameworks like \textbf{$KG^2RAG$} \cite{KG2RAG} adopt a "Semantic-to-Graph" paradigm, using text seeds from initial semantic retrieval to expand graph components. However, these methods often face a "unidirectional bottleneck," where the strengths of one modality do not actively feed back to correct or complete the other in real-time. 

To overcome these limitations, recent frameworks have moved toward cross-source validation. 
\textbf{Think-on-Graph 2.0 (ToG-2)} \cite{ICLR2025_830b1abc} introduces a tight-coupling paradigm that iteratively alternates between KG-based graph retrieval and document-based context retrieval, using text evidence to prune graph paths and graph structures to guide text retrieval. 
While TGS-RAG shares the high-level goal of mutual verification, the underlying mechanisms differ substantially. 
ToG-2 follows a subtractive strategy, repeatedly pruning and re-expanding the search space through costly iterative retrieval. 
In contrast, TGS-RAG adopts an additive strategy via Memory-based Orphan Entity Bridging, which treats pruning as postponement rather than elimination and directly resurrects pruned paths from the semantic beam search history. 
This retrospective repair mechanism enables efficient recovery of logical chains that are difficult to preserve under sparse graph connectivity and search-time pruning without additional database queries.

\begin{figure*}[t]
    \centering
    \includegraphics[width=\textwidth]{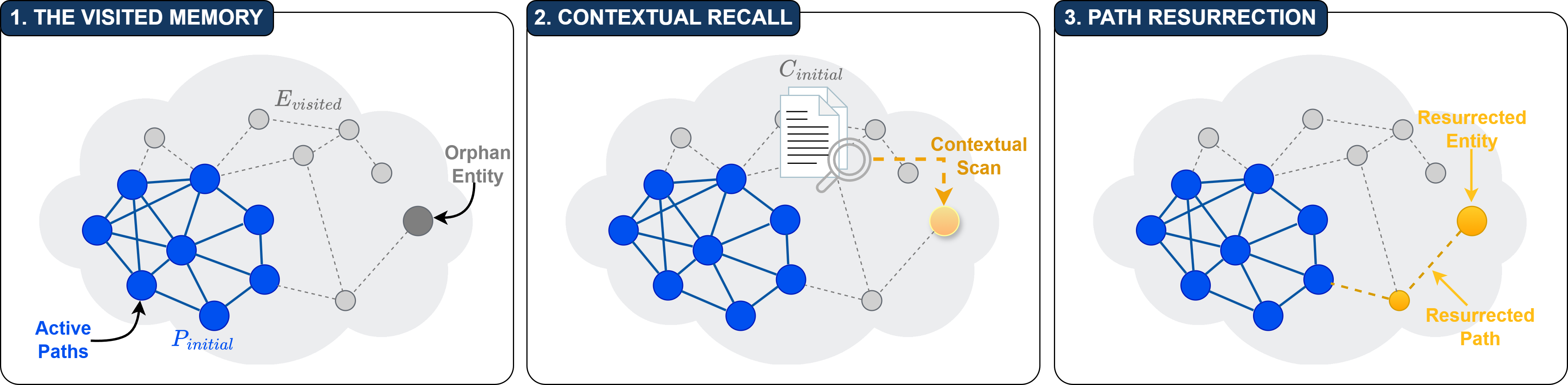}
    \caption{Illustration of the Memory-based Orphan Entity Bridging algorithm. The process operates in three steps: (1) \textbf{The Visited Memory}: Semantic Beam Search generates a set of \textit{Active Paths} ($P_{initial}$, blue nodes) while implicitly storing pruned nodes in the \textit{Visited Memory} ($E_{visited}$, grey area); (2) \textbf{Contextual Recall}: Contextual clues from retrieved text chunks ($C_{initial}$) perform a \textbf{Contextual Scan} to identify relevant but pruned "orphan entities"; (3) \textbf{Path Resurrection}: The pruning decision is reversed, and the \textit{Resurrected Path} ($p_{bridge}$, orange) connecting the resurrected entity to the main subgraph is recovered without additional database queries.}
    \label{fig:bridging}
\end{figure*}

\section{The TGS-RAG Framework}

As illustrated in Figure \ref{fig:framework}, TGS-RAG is designed as a two-stage framework that tightly integrates unstructured text and structured knowledge graphs for retrieval-augmented reasoning.  It consists of an offline \textbf{Knowledge Base Construction} stage, responsible for extracting and fusing knowledge from unstructured documents; and an online \textbf{Bidirectional Synergistic Retrieval} stage, which performs mutual verification and enhancement between text and graph evidence during inference.

\subsection{Knowledge Base Construction}

Given an unstructured document corpus $C$, TGS-RAG constructs a unified knowledge base through an offline processing pipeline. We leverage Large Language Models (LLMs) to extract salient entities $E$ and their semantic relations $R$ (see extraction prompt in Appendix A.1), forming a global knowledge graph $G=(E, R)$.

Crucially, we establish a \textbf{bidirectional mapping} $M$ between the text corpus $C$ and the knowledge graph $G$. Each relational triplet in $G$ is explicitly linked to its originating text chunks, while each text chunk is annotated with the entities and relations it contains. This mapping transforms the knowledge graph from a standalone structure into a reasoning interface over text. Formally, the knowledge base is defined as a triplet:

\begin{equation}
    KB = (C, G, M)
\end{equation}
where $M$ represents the bidirectional links between $C$ and $G$. This unified and interconnected knowledge base serves as the foundation for the proposed bidirectional enhanced retrieval framework.

\subsection{Dual-Channel Initial Retrieval}

Given a user query $q$, TGS-RAG first extracts a set of query-relevant entity mentions using an LLM. These entities are embedded to obtain vectors $v_E$, which are then used to retrieve a seed entity set $E_{\text{seed}}$ from the knowledge graph $G$.

Starting from $E_{\text{seed}}$, TGS-RAG performs two complementary retrieval processes in parallel: a Text Channel and a Graph Channel. This dual-channel design intentionally captures heterogeneous evidence signals that exhibit different failure modes.

\begin{itemize}
    \item \textbf{Text Channel}: This channel performs standard semantic retrieval to quickly recall textual evidence directly related to the user query. We use the query vector $v_q$ of user query $q$ to perform vector indexing on the text chunk collection $C$, retrieving the initial text set $C_{initial}$ with the highest similarity to $v_q$.
    \item \textbf{Graph Channel}: Starting from the seed entity set $E_{seed}$, the Semantic Beam Search algorithm (see Algorithm 1 in Appendix B) computes, at each hop, the cosine similarity between the embeddings of neighboring nodes and the query vector $v_q$. Only the top-$K$ (Beam Width) most semantically relevant paths are retained for the next expansion step. This heuristic pruning ensures that the retrieval focuses on the semantic neighborhood of the query, yielding a high-quality path set $P_{initial}$ within a preset depth $d$.
\end{itemize}

The outcome of this stage is two heterogeneous evidence sets: $C_{\text{initial}}$, which emphasizes semantic recall, and $P_{\text{initial}}$, which captures structured reasoning paths.

\subsection{Synergistic Fusion and Enhancement}

After obtaining the initial text evidence $C_{initial}$ and graph path set $P_{initial}$, the TGS-RAG introduces a bidirectional scoring and discovery mechanism that enables mutual verification and refinement of information. This process includes two enhancement directions: \emph{Graph-to-Text} and \emph{Text-to-Graph}.

\subsubsection{Graph-to-Text Synergy}
This process utilizes structured knowledge from the graph channel to guide the rediscovery and re-ranking of text evidence, rather than relying only on surface-level similarity to the query.

\begin{enumerate}
    \item \textbf{Chunk Recommendation}: We aggregate all entities visited during the Semantic Beam Search (including those in pruned paths) to form a global visited set $E_{visited}$. These entities act as "recommenders" to "vote" for their source text chunks. The graph-recommended chunks are merged with $C_{initial}$ to create a broader candidate pool $C_{candidate}$, allowing text chunks that have low semantic similarity but are structurally relevant to be recalled.
    \item \textbf{Chunk Re-ranking}: We calculate a recommendation score $Rec(c)$ for each text chunk $c \in C_{candidate}$, proportional to the number of entities recommending it. The final score $Score_{final}(c)$ is a weighted fusion of its original semantic similarity and the graph recommendation score:
    \begin{align}
        Score_{\text{final}}(c) =\; & \alpha \cdot \text{Norm}\bigl(\text{sim}(v_q, v_c)\bigr) \nonumber \\
        & + (1 - \alpha) \cdot \text{Norm}\bigl(\text{Rec}(c)\bigr)
    \end{align}
    where $\alpha$ is a hyperparameter balancing the two scores.
\end{enumerate}

\begin{table*}[t]
\centering
\resizebox{\textwidth}{!}{%
\begin{tabular}{ll|cccc|c}
\toprule
\multirow{2}{*}{\textbf{Dataset}} & \multirow{2}{*}{\textbf{Method}} & \multicolumn{4}{c|}{\textbf{Retrieval Metrics}} & \multicolumn{1}{c}{\textbf{Generation}} \\
\cmidrule(lr){3-6} \cmidrule(lr){7-7}
 & & \textbf{Strict Hit Rate (\%)} & \textbf{Recall (\%)} & \textbf{Precision (\%)} & \textbf{Support F1 (\%)} & \textbf{LLM Judge Acc (\%)} \\
\midrule
\multirow{6}{*}{\textbf{MuSiQue}} 
 & Naive RAG & 14.23 & 43.96 & 21.17 & 28.11 & 21.51 \\
 & Hybrid RAG & 14.23 & 43.98 & 21.18 & 28.13 & 21.56 \\
 & GraphRAG (Local) & 30.70 & 59.62 & 7.19 & 12.59 & 40.67 \\
 & LightRAG (Max) & 33.54 & 60.44 & 20.23 & 18.58 & 40.81 \\
 & $KG^2RAG$ & 9.10 & 31.02 & 11.50 & 15.61 & 15.06 \\
 \cmidrule(lr){2-7}
 & \textbf{TGS-RAG (Ours)} & \textbf{34.84} & \textbf{62.01} & \textbf{24.85} & \textbf{20.60} & \textbf{41.37} \\
\midrule 
\multirow{6}{*}{\textbf{HotpotQA}} 
 & Naive RAG & 45.01 & 62.48 & 26.99 & 38.56 & 59.41 \\
 & Hybrid RAG & 46.05 & 65.45 & 27.62 & 39.42 & 61.92 \\
 & GraphRAG (Local) & 55.78 & 70.15 & 10.11 & 20.85 & 71.79 \\
 & LightRAG (Max) & 60.55 & 72.04 & 10.23 & 15.58 & 72.09 \\
 & $KG^2RAG$ & 38.83 & 58.92 & 16.49 & 25.40 & 60.46 \\
 \cmidrule(lr){2-7}
 & \textbf{TGS-RAG} & \textbf{62.00} & \textbf{77.55} & \textbf{27.41} & \textbf{26.06} & \textbf{79.99} \\
\bottomrule
\end{tabular}%
}
\caption{Main results on MuSiQue and HotpotQA datasets. Retrieval performance is evaluated using Provenance Mapping. Generation quality is evaluated by DeepSeek-V3.2 as a Judge.}
\label{tab:main_results}
\end{table*}

\subsubsection{Text-to-Graph Synergy}

This process utilizes the rich contextual information in the text channel to verify and enhance the structured paths discovered in the graph channel. As illustrated in Figure \ref{fig:bridging}, this involves confirming existing paths and resurrecting pruned connections.

For each path $p \in P_{initial}$, we first calculate a base quality score $Score_{base}(p)$ based on the semantic similarity of its components to the query $q$ and their structural importance.

\begin{enumerate}
    \item \textbf{Path Confirmation}: The path is validated by checking the intersection of entities on the path and entities contained in $C_{initial}$. If a path receives support from text evidence (i.e., its entities appear in the retrieved text chunks), its score is boosted:
    \begin{multline}
        Score_{conf}(p) = Score_{base}(p) \\
        + \epsilon \cdot |Entities(p) \cap Entities(C_{initial})|
    \end{multline}
    where $\epsilon$ is the weight for text confirmation reward.
    
    \item \textbf{Memory-based Orphan Entity Bridging}: To discover potential knowledge ignored by the initial path pruning, we define \textbf{Orphan Entities} $E_{orphan}$ as key entities that are activated by the retrieved text evidence $C_{initial}$ 
but are absent from the initially selected graph paths $P_{initial}$. Instead of performing a costly new search or re-expansion, we leverage the \textbf{Visited Memory} as a 
\emph{deferred reasoning buffer}. If an orphan entity exists in the visited but pruned nodes, 
the original pruning decision is reversed by directly replaying its stored path, 
thereby resurrecting a valid reasoning chain without additional database queries. We check if any orphan entity exists in the set of visited nodes that were pruned during the beam selection (detailed procedure in Algorithm 2, Appendix B). If found, the path leading to this orphan entity is "resurrected" as a bridge path $p_{bridge}$. This mechanism efficiently recovers logical connections that are semantically distant but contextually bridged by the text, without incurring additional database query overhead.
\end{enumerate}

\subsection{Context Consolidation and Answer Generation}

After precise scoring and ranking, the Top-K paths $P_{Top-K}$ and text chunks $C_{Top-K}$ are selected. Each $P_{Top-K}$ is formatted into clear natural language reasoning chains. If supported by $p_{bridge}$, these bridging paths are also appended as supporting metadata. Finally, the consolidated context of $P_{Top-K}$ and $C_{Top-K}$, along with the original query $q$, is submitted to the LLM to generate a logically coherent and evidence-grounded answer.

\section{Experiments}
\label{sec:experiments}

In this section, we present a comprehensive evaluation of TGS-RAG against state-of-the-art retrieval paradigms. We aim to validate the effectiveness of our bidirectional synergy mechanism in resolving the "information island" problem inherent in multi-hop reasoning tasks.

\subsection{Experimental Setup}

\subsubsection{Datasets}
We utilize two widely recognized benchmarks for multi-hop question answering:
\begin{itemize}
    \item \textbf{MuSiQue-Ans} \cite{trivedi2021musique}: A dataset characterized by connected reasoning chains (typically 2-4 hops) and low lexical overlap between connected documents. It serves as the primary testbed for our "Orphan Entity Bridging" mechanism.
    \item \textbf{HotpotQA (Distractor)} \cite{yang-etal-2018-hotpotqa}: A dataset requiring reasoning over two supporting documents amidst logically irrelevant but semantically similar distractor documents.
\end{itemize}

\subsubsection{Implementation Details}
To ensure a robust and cost-effective experimental environment, we employ \textbf{GPT-4o-mini} as the backbone LLM for all tasks, including Knowledge Graph construction, entity extraction, and final answer generation. 
For semantic representation, we utilize \textbf{Qwen3-Embedding-0.6B}\citep{zhang2025qwen3embeddingadvancingtext} as the unified embedding model for all queries, text chunks, and graph entities.
For evaluation, we utilize DeepSeek-V3.2 \citep{liu2024deepseek} as an impartial LLM-as-a-Judge \citep{li2025generation} (evaluation prompt provided in Appendix A.4). We instruct DeepSeek-V3.2 to evaluate semantic equivalence between the generated answer and the ground truth, accounting for valid aliases while penalizing hallucinations.

\begin{table*}[t]
\centering
\resizebox{0.8\textwidth}{!}{%
\begin{tabular}{ll|rrrr}
\toprule
\multirow{2}{*}{\textbf{Dataset}} & \multirow{2}{*}{\textbf{Method}} & \multicolumn{4}{c}{\textbf{Token Usage}} \\
\cmidrule(lr){3-6}
& & \textbf{Embedding} & \textbf{LLM Prompt} & \textbf{LLM Completion} & \textbf{Total Cost} \\
\midrule
\multirow{6}{*}{\textbf{MuSiQue}} 
 & Naive RAG & 54,735 & 1,756,984 & 13,461 & 1,825,180 \\
 & Hybrid RAG & 54,735 & 1,758,189 & 13,437 & 1,826,361 \\
 & GraphRAG (Local) & 51,845 & 13,908,681 & 554,681 & 14,515,224 \\
 & LightRAG (Max) & 174,024 & 16,541,984 & 374,635 & 17,090,643 \\
 & $KG^2RAG$ & 1,795,495 & 909,001 & 10,213 & 2,714,709 \\
 \cmidrule(lr){2-6}
 & \textbf{TGS-RAG} & \textbf{78,030} & \textbf{5,113,149} & \textbf{252,224} & \textbf{5,443,403} \\
\midrule
\multirow{6}{*}{\textbf{HotpotQA}} 
 & Naive RAG & 2,202,470 & 59,485,904 & 345,673 & 62,034,047 \\
 & Hybrid RAG & 2,280,240 & 63,283,087 & 346,239 & 65,909,566 \\
 & GraphRAG (Local) & 1,475,556 & 634,239,236 & 10,674,261 & 646,389,053 \\
 & LightRAG (Max) & 4,574,224 & 746,163,808 & 7,163,934 & 757,901,966 \\
 & $KG^2RAG$ & 2,119,623 & 41,072,699 & 398,399 & 43,590,721 \\
 \cmidrule(lr){2-6}
 & \textbf{TGS-RAG} & \textbf{3,406,165} & \textbf{187,692,537} & \textbf{26,438,352} & \textbf{217,537,053} \\
\bottomrule
\end{tabular}%
}
\caption{Efficiency and token usage analysis. Detailed breakdown of token consumption across MuSiQue and HotpotQA datasets. Grouping by dataset allows for a direct comparison of computational costs among different retrieval paradigms.}
\label{tab:efficiency_token}
\end{table*}

\subsubsection{Baselines}
We compare TGS-RAG with three categories of retrieval methods. All baselines are implemented following their original configurations, utilizing identical LLM backbones (GPT-4o-mini) and consistent text chunking strategies to ensure a fair comparison. Detailed implementation parameters and sensitivity analyses are provided in Appendix C and E.

\begin{itemize}
    \item \textbf{Naive RAG:} Standard dense retrieval using vector similarity.
    \item \textbf{Hybrid RAG (Dense+Sparse Text Retrieval):} A combination of dense vector retrieval and sparse keyword retrieval (BM25/TF-IDF) fused via Reciprocal Rank Fusion (RRF) \cite{10.1145/1571941.1572114}.
    \item \textbf{Graph-centric Methods:} We evaluate state-of-the-art graph retrieval systems, including \textbf{GraphRAG} \citep{edge2024local} and \textbf{LightRAG} \citep{guo-etal-2025-lightrag}. To rigorously benchmark the performance ceiling of each method, we employ their most comprehensive retrieval modes and include their associated textual evidence or community summaries where applicable: "Local Search" for GraphRAG (focusing on entity-centric reasoning) and "Max" mode for LightRAG (integrating both global and local signals). Additionally, we compare against \textbf{$KG^2RAG$} \citep{KG2RAG} as a representative of semantic-to-graph expansion paradigms.
\end{itemize}

For methods with explicit text chunk retrieval, we fix the text chunk top-$k$ to $k=5$.

\subsubsection{Evaluation Metrics}
To rigorously assess both retrieval completeness and generation accuracy, we employ a multi-dimensional metric system.

\paragraph{Retrieval Scope with Provenance Mapping.}
Unlike standard RAG which only retrieves text chunks, TGS-RAG retrieves both structured paths and unstructured chunks. To ensure a fair comparison with document-level ground truth, we define the set of retrieved documents $D_{ret}$ as the union of explicit document retrieval and implicit graph provenance:
\begin{equation}
    D_{ret} = D_{chunks} \cup \{doc \mid \exists e \in P_{graph}, doc \in Source(e)\}
\end{equation}
where $D_{chunks}$ denotes the source documents of the top-$k$ retrieved text chunks, and $Source(e)$ represents the source documents of entities contained in the retrieved graph paths $P_{graph}$.
Based on $D_{ret}$, we calculate \textbf{Strict Hit Rate (SHR)}, the percentage of queries where $D_{ret}$ contains \textit{all} ground truth supporting documents, and \textbf{Support F1}, measuring the quality of evidence. Because graph-based retrieval expands evidence through entity-to-document provenance, its retrieved document set is typically broader than chunk-only baselines, making Support F1 a conservative metric that may be lower despite stronger structural coverage.

\paragraph{Generation Quality (LLM-as-a-Judge).}
Traditional string-matching metrics (e.g., Exact Match) often penalize correct but verbose answers. Therefore, we adopt \textbf{LLM Judge Accuracy}, employing DeepSeek-V3.2 to judge whether the generated response contains the correct information specified in the gold answer.

\subsection{Main Results}

\subsubsection{Performance Comparison}
Table \ref{tab:main_results} presents a comprehensive evaluation of retrieval and generation performance across MuSiQue and HotpotQA benchmarks.

\paragraph{Limitations of Shallow Hybridization and Unidirectional Expansion.}
The results reveal critical bottlenecks in existing retrieval paradigms. As observed in the MuSiQue dataset, \textbf{Hybrid RAG} yields a Strict Hit Rate (14.23\%) and Judge Accuracy (21.56\%) virtually identical to Naive RAG. This stagnation confirms that the primary challenge in multi-hop reasoning is not lexical mismatch—which keyword search addresses—but \textit{structural disconnection}. Hybrid methods fail to bridge the gap when documents share no lexical overlap but are logically linked. Similarly, \textbf{$KG^2RAG$} underperforms significantly (9.10\% Hit Rate on MuSiQue), demonstrating the fragility of unidirectional "semantic-to-graph" expansion; if the initial semantic retrieval misses key anchor points, the subsequent graph reasoning collapses due to the lack of a feedback loop.

\paragraph{The Precision-Recall Trade-off in Graph Baselines.}
While heavy graph-based approaches like \textbf{GraphRAG} and \textbf{LightRAG} achieve competitive recall, they suffer from severe precision degradation. On HotpotQA, both methods exhibit extremely low retrieval precision ($\approx$10\%), compared to 27.41\% for TGS-RAG. This indicates an \textit{information overload} phenomenon: these systems retrieve excessive, logically loosely related graph components (e.g., entire communities), which introduces significant noise into the context window. Consequently, despite high recall, their generation accuracy (LLM Judge) is compromised by the distraction of irrelevant evidence.

\paragraph{Superiority of TGS-RAG.}
\textbf{TGS-RAG} establishes a new state-of-the-art by effectively balancing retrieval scope and precision. On MuSiQue, it achieves a Strict Hit Rate of \textbf{34.84\%}, more than doubling the performance of Hybrid RAG. On HotpotQA, it attains the highest Strict Hit Rate, Recall, and Judge Accuracy, reaching a Judge Accuracy of \textbf{79.99\%} despite a conservative Support F1 of \textbf{26.06\%} under broader graph provenance. Unlike baselines that trade precision for recall, TGS-RAG utilizes the bidirectional synergy to filter out pseudo-evidence while resurrecting subtle logical links, providing the LLM with a concise and rigorous reasoning chain.

\subsection{Efficiency Analysis}
Table \ref{tab:efficiency_token} details the computational costs associated with each paradigm. A comparative analysis highlights the superior cost-effectiveness of our framework.

\paragraph{Prohibitive Costs of Global Indexing.}
Graph-based global indexing methods impose a prohibitive computational burden. \textbf{LightRAG} and \textbf{GraphRAG} incur astronomical token usage (e.g., over 757M tokens for LightRAG on HotpotQA vs. 62M for Naive RAG), primarily driven by the exhaustive construction of community summaries and dual-level indices. This massive overhead renders them impractical for dynamic or large-scale knowledge bases where frequent updates are required.

\paragraph{Optimal Pareto Frontier of TGS-RAG.}
TGS-RAG achieves an optimal trade-off between performance and efficiency. While it incurs a moderate cost increase over Naive RAG ($\approx$3x) to support graph reasoning, it is drastically more efficient than global graph methods—consuming only \textbf{37\%} of the tokens used by GraphRAG on MuSiQue and less than \textbf{30\%} of LightRAG on HotpotQA. This efficiency stems from our algorithmic design: instead of pre-computing expensive global summaries, TGS-RAG employs an \textit{on-demand} retrieval strategy. The \textit{Semantic Beam Search} and \textit{Memory-based Orphan Entity Bridging} selectively explore and resurrect only contextually relevant paths during inference, avoiding the redundant processing of the entire graph structure.

\subsection{Ablation Study}

To verify the contribution of each module in our bidirectional synergy mechanism, we conducted an ablation study on the \textbf{HotpotQA} dataset. We developed two variants:
\begin{itemize}
    \item \textbf{w/o Graph-to-Text (Re-ranking):} Disables the \textit{Global Voting} mechanism. Text chunks are ranked solely by vector similarity.
    \item \textbf{w/o Text-to-Graph (Bridging):} Disables the \textit{Memory-based Orphan Entity Bridging}. The system relies solely on the initial graph paths found by beam search.
\end{itemize}

\begin{table}[t]
\centering
\resizebox{\linewidth}{!}{%
\begin{tabular}{l|cccc}
\toprule
\textbf{Variant} & \textbf{SHR (\%)} & \textbf{Recall (\%)} & \textbf{Prec (\%)} & \textbf{F1 (\%)} \\
\midrule
\textbf{TGS-RAG (Full)} & \textbf{62.00} & \textbf{77.55} & \textbf{27.41} & \textbf{26.06} \\
w/o Bridging & 47.82 & 68.16 & 22.74 & 23.46 \\
w/o Re-ranking & 52.65 & 70.31 & 23.88 & 24.08 \\
\bottomrule
\end{tabular}%
}
\caption{Ablation study on HotpotQA. Detailed impact of removing synergy modules.}
\label{tab:ablation}
\end{table}

As shown in Table \ref{tab:ablation}, removing the Graph-Guided Re-ranking module leads to a decline in Strict Hit Rate to \textbf{52.65\%} and Support F1 to \textbf{24.08\%}. This indicates that without structured filtering, the retriever struggles to distinguish between semantically similar distractors and true supporting facts, thereby reducing the precision and overall quality of the evidence context. A concrete example of this "Semantic Drift" and how our synergy mechanism resolves it is analyzed in Appendix D.

Similarly,  removing the Bridging module results in the most significant performance degradation, with Strict Hit Rate dropping sharply from \textbf{62.00\%} to \textbf{47.82\%}. This confirms that standard graph traversal may fail when potentially valid evidence nodes are pruned early, whereas our memory-based resurrection mechanism effectively recovers these "orphan" connections.

\section{Conclusion}

We have proposed \textbf{TGS-RAG}, a bidirectional text–graph synergistic framework designed to resolve the isolation between unstructured textual evidence and structured graph-based reasoning. By modeling retrieval as a collaborative process, TGS-RAG effectively bridges the "Information Island" gap inherent in multi-hop reasoning tasks.

The framework introduces two complementary enhancement channels: \textbf{Graph-Guided Re-ranking} leverages structured logic to filter semantic noise from text retrieval, while \textbf{Memory-based Orphan Entity Bridging} utilizes textual context to validate and complete graph reasoning paths. A critical innovation is the use of "visited memory" to \textit{resurrect} pruned paths, which recovers potentially useful logical links that were previously discarded during search, without the computational overhead of iterative expansion.

Extensive experiments on MuSiQue and HotpotQA demonstrate that TGS-RAG consistently outperforms strong text-based, graph-based, and hybrid baselines. Our results indicate that shallow hybridization is insufficient; instead, effective multi-hop reasoning requires structured and unstructured evidence to mutually verify and correct each other. Notably, TGS-RAG achieves this state-of-the-art performance with significantly lower token consumption compared to global graph retrieval methods. Overall, this work underscores the importance of bidirectional synergy and provides a scalable foundation for integrating symbolic structure with neural representations in LLMs.

\clearpage
\appendix
\input{Appendix}

\bibliographystyle{named}
\bibliography{ijcai26}

\end{document}

%% file: Appendix.tex
\section{Prompt Templates}
\label{app:prompts}

To ensure reproducibility, we provide the core prompt templates used in \tgsrag. All prompts are engineered to be concise and task-specific for the underlying LLM (GPT-4o-mini).

\subsection{Entity \& Relation Extraction Prompt}
This prompt guides the LLM to extract structured knowledge triplets from unstructured text chunks during the Knowledge Base Construction phase.

\begin{tcolorbox}[breakable, enhanced, colback=gray!5, colframe=gray!40, title=\textbf{System Prompt: Knowledge Graph Specialist}, fonttitle=\bfseries, fontupper=\scriptsize, boxsep=1mm, left=1mm, right=1mm, top=1mm, bottom=1mm, before skip=4pt, after skip=4pt]
\textbf{---Role---}\\
You are a Knowledge Graph Specialist responsible for extracting entities and relationships from the input text.

\textbf{---Instructions---}
\begin{enumerate}[leftmargin=*, nosep]
    \item \textbf{Entity Extraction:} Identify clearly defined and meaningful entities.
    \begin{itemize}[nosep]
        \item \textbf{Fields:}
        \begin{itemize}
             \item \texttt{name}: The name of the entity (Title Case).
             \item \texttt{type}: Categorize the entity using the provided \texttt{Entity\_types}. If none apply, classify as \texttt{other}.
             \item \texttt{description}: A concise description based \textit{solely} on the text.
        \end{itemize}
        \item \textbf{Target Format (JSON Lines):}\\
        \texttt{\{"type": "entity", "name": "...", "category": "...", "desc": "..."\}}
    \end{itemize}

    \item \textbf{Relationship Extraction:} Identify direct binary relationships between previously extracted entities.
    \begin{itemize}[nosep]
        \item \textbf{Rule:} Decompose complex N-ary relationships into binary pairs.
        \item \textbf{Fields:}
        \begin{itemize}
             \item \texttt{source} / \texttt{target}: The exact names of the source and target entities.
             \item \texttt{keywords}: Comma-separated high-level keywords summarizing the relation.
             \item \texttt{description}: A concise explanation of the connection.
        \end{itemize}
        \item \textbf{Target Format (JSON Lines):}\\
        \texttt{\{"type": "relation", "source": "...", "target": "...", "keywords": ["..."], "desc": "..."\}}
    \end{itemize}

    \item \textbf{General Rules:}
    \begin{itemize}[nosep]
        \item Output all entities first, followed by all relationships.
        \item Use third-person perspective. Avoid pronouns like 'I', 'you', 'this article'.
        \item The entire output must be in \texttt{\{language\}}. Proper nouns should be retained in their original language.
        \item \textbf{End Signal:} Output the literal string \texttt{<|COMPLETE|>} on the final line.
    \end{itemize}
\end{enumerate}

\textbf{---Real Data to be Processed---}\\
<Input>\\
Entity\_types: \texttt{[\{entity\_types\}]}\\
Text:\\
\texttt{\{input\_text\}}
\end{tcolorbox}

\subsection{Query Entity Extraction Prompt}
This prompt is used in the online retrieval stage. To handle complex user intents and avoid keyword noise, we employ detailed exclusion rules and few-shot demonstrations to guide the LLM in identifying the core search targets.

\begin{tcolorbox}[breakable, enhanced, colback=gray!5, colframe=gray!40, title=\textbf{User Prompt: Query Analysis}, fonttitle=\bfseries, fontupper=\scriptsize, boxsep=1mm, left=1mm, right=1mm, top=1mm, bottom=1mm, before skip=4pt, after skip=4pt]
\textbf{---Role---}\\
You are a highly intelligent query analysis engine for a Retrieval-Augmented Generation (RAG) system.

\textbf{---Goal---}\\
Accurately identify and extract key concepts or entities that serve as the \textbf{core subjects} of the user's query.

\textbf{---Definition of Core Entity---}\\
A core entity is a noun or proper noun with clear referential meaning. It usually belongs to:
\begin{itemize}[nosep]
    \item \textbf{Specific Objects:} Person names, locations, organizations, models (e.g., "iPhone 15").
    \item \textbf{Abstract Concepts:} Technical terms, theories, strategies (e.g., "RoHS directive").
    \item \textbf{Broad Themes:} The central topic of vague queries (e.g., extract "marketing" from "tell me about marketing").
\end{itemize}

\textbf{---Exclusion Criteria (Do NOT Extract)---}\\
1. \textbf{User Intent Verbs:} Words indicating what the user wants to do (e.g., "compare", "find", "list", "describe").\\
2. \textbf{Functional Words:} Stop words or general nouns acting as sentence structures (e.g., "information", "detail", "introduction").

\textbf{---Few-Shot Demonstrations---}
\begin{enumerate}[leftmargin=*, nosep]
    \item \textit{User:} "Help me introduce a fighter."\\
          \textit{Output:} \texttt{["fighter"]} (Ignore "introduce")
    \item \textit{User:} "What is the relationship between RoHS and peak forward current?"\\
          \textit{Output:} \texttt{["RoHS", "peak forward current"]}
    \item \textit{User:} "Hello, how are you?"\\
          \textit{Output:} \texttt{[]} (No core entities)
\end{enumerate}

\textbf{---Task---}\\
User Query: \texttt{"[User Query String]"}\\
Extracted Entities (JSON Format):
\end{tcolorbox}

\subsection{Final Answer Generation Prompt}
This prompt synthesizes the retrieved graph paths and text chunks. It includes strict instructions to prioritize evidence over prior knowledge and explicitly handle missing information to prevent hallucinations.

\begin{tcolorbox}[breakable, enhanced, colback=gray!5, colframe=gray!40, title=\textbf{User Prompt: Context Consolidation}, fonttitle=\bfseries, fontupper=\scriptsize, boxsep=1mm, left=1mm, right=1mm, top=1mm, bottom=1mm, before skip=4pt, after skip=4pt]
\textbf{---Role---}\\
You are a knowledgeable and logically rigorous AI knowledge assistant.

\textbf{---Core Instructions---}
\begin{enumerate}[leftmargin=*, nosep]
    \item \textbf{Strict Adherence to Context:} Your answer must be \textbf{completely} and \textbf{solely} based on the provided evidence. You are \textbf{strictly prohibited} from using internal prior knowledge.
    \item \textbf{Synthesis \& Reasoning:} 
    \begin{itemize}[nosep]
        \item Use \textbf{Knowledge Graph Paths} to establish the logical backbone (relationships between A and B).
        \item Use \textbf{Textual Evidence} to fill in specific details (dates, descriptions, attributes).
        \item If a logical link is provided in the graph but missing in the text (or vice versa), synthesize them to form a complete chain.
    \end{itemize}
    \item \textbf{Anti-Hallucination:} If the provided context does not contain sufficient information to answer the question, explicitly state: \textit{"Based on the provided materials, I cannot answer this question."}
    \item \textbf{Structure:} Organize the response clearly using headings, bullet points, or Markdown tables for comparisons.
\end{enumerate}

\textbf{---Context Provided---}\\
\textbf{\#\# Knowledge Graph Paths}\\
\texttt{[Path 1: Entity A --(relation)--> Entity B --(relation)--> Entity C]}\\
\texttt{[Path 2: ...]}

\textbf{\#\# Textual Evidence}\\
\texttt{[Evidence 1 (Source: Doc A): "...specific content text..."]}\\
\texttt{[Evidence 2: ...]}

\textbf{---Task---}\\
User's Original Question: \texttt{"[User Query String]"}\\
Your Answer:
\end{tcolorbox}

\subsection{LLM-as-a-Judge Evaluation Prompt}
\label{app:judge}

To evaluate generation quality, we adopt an LLM-as-a-Judge paradigm to assess whether a generated answer contains the correct information specified by the gold answer.

\begin{tcolorbox}[breakable, enhanced, colback=gray!5, colframe=gray!40, title=\textbf{System Prompt: Answer Correctness Judge}, fonttitle=\bfseries, fontupper=\scriptsize, boxsep=1mm, left=1mm, right=1mm, top=1mm, bottom=1mm, before skip=4pt, after skip=4pt]
You are an impartial and strict judge evaluating the correctness of a generated answer compared to a gold standard answer. Your task is to determine if the "Generated Answer" contains the correct information specified in the "Gold Answer".

\textbf{Rules:}
\begin{enumerate}[leftmargin=*, nosep]
    \item \textbf{Answer-Centric Evaluation:} Judge solely based on whether the core answer required by the Gold Answer is correctly provided. Additional correct or irrelevant information should not affect the judgement unless it introduces contradictions.
    \item \textbf{Semantic Equivalence:} If the generated answer is verbose but clearly contains the correct core entity, fact, or conclusion, mark it as \texttt{CORRECT}.
    \item \textbf{Aliases:} Treat any provided aliases of the Gold Answer as equally valid correct answers.
    \item \textbf{Hallucination or Contradiction:} If the generated answer contains incorrect facts or conflicts with the Gold Answer, mark it as \texttt{INCORRECT}.
    \item \textbf{Non-Answering:} If the generated answer states uncertainty or fails to provide the required answer, mark it as \texttt{INCORRECT}.
\end{enumerate}

\textbf{Output Format (JSON):}
\begin{verbatim}
{
  "is_correct": boolean,
  "reason": "Short explanation."
}
\end{verbatim}
\end{tcolorbox}

\section{Algorithm Pseudocode}
\label{app:algo}

We present the pseudocode for the core algorithms of \tgsrag: Semantic Beam Search (which constructs the Visited Memory) and Memory-based Orphan Entity Bridging.

\begin{algorithm}[H]
\caption{Semantic Beam Search with Visited Memory}
\label{alg:beam_search}
\begin{algorithmic}[1]
\REQUIRE Query embedding $\mathbf{q}$, Seed Entities $E_{seed}$, Knowledge Graph $\mathcal{G}$, Beam Width $K$, Max Depth $D$
\ENSURE Top Paths $P_{initial}$, Visited Memory $\mathcal{M}_{visited}$
\STATE \textbf{Initialize} $P_{curr} \leftarrow \{[e] \mid e \in E_{seed}\}$
\STATE \textbf{Initialize} $\mathcal{M}_{visited} \leftarrow \emptyset$ 
\FOR{each $e \in E_{seed}$}
    \STATE $\mathcal{M}_{visited}[e] \leftarrow \{path: [e], score: 1.0\}$ \COMMENT{Initialize memory with seeds}
\ENDFOR

\FOR{$d = 1$ to $D$}
    \STATE $Candidates \leftarrow \emptyset$
    \FOR{each path $p \in P_{curr}$}
        \STATE $e_{curr} \leftarrow p_{last}$ \COMMENT{Last node in path}
        \STATE $\mathcal{N} \leftarrow \mathcal{G}.\text{GetNeighbors}(e_{curr})$ \COMMENT{Limit by edge weight if necessary}
        \FOR{each neighbor $n \in \mathcal{N}$}
            \IF[Cycle prevention]{$n \notin p$}
                \STATE $\mathbf{n} \leftarrow \text{GetEmbedding}(n)$
                \STATE $sim \leftarrow \text{CosineSim}(\mathbf{n}, \mathbf{q})$
                \STATE $p_{new} \leftarrow p + [n]$ \COMMENT{\textbf{Key Step:} Track all explored nodes and their paths in Memory}
                \IF{$n \notin \mathcal{M}_{visited}$ \textbf{or} $sim > \mathcal{M}_{visited}[n].score$}
                    \STATE $\mathcal{M}_{visited}[n] \leftarrow \{path: p_{new}, score: sim\}$
                \ENDIF
                \STATE $Candidates \leftarrow Candidates \cup \{(p_{new}, sim)\}$
            \ENDIF
        \ENDFOR
    \ENDFOR
    \STATE $P_{curr} \leftarrow \text{SelectTopK}(Candidates, K)$ \COMMENT{Pruning step: keep only top-K paths}
\ENDFOR
\RETURN $P_{curr}, \mathcal{M}_{visited}$
\end{algorithmic}
\end{algorithm}

\begin{algorithm}[H]
\caption{Memory-based Orphan Entity Bridging}
\label{alg:bridging}
\begin{algorithmic}[1]
\REQUIRE Retrieved Text Chunks $C_{text}$, Visited Memory $\mathcal{M}_{visited}$, Initial Graph Paths $P_{initial}$
\ENSURE Final Path Set $P_{final}$
\STATE $E_{text} \leftarrow \text{ExtractEntityIds}(C_{text})$
\STATE $E_{graph} \leftarrow \bigcup_{p \in P_{initial}} \{e \mid e \in p\}$
\STATE $E_{orphan} \leftarrow E_{text} \setminus E_{graph}$ \COMMENT{Identify entities present in text but missing from graph paths}
\STATE $P_{bridge} \leftarrow \emptyset$

\FOR{each orphan entity $e \in E_{orphan}$}
    \IF{$e \in \mathcal{M}_{visited}$}
        \STATE $p_{recovered} \leftarrow \mathcal{M}_{visited}[e].path$ \COMMENT{Zero-overhead bridging: Resurrect path from memory}
        \STATE $P_{bridge} \leftarrow P_{bridge} \cup \{p_{recovered}\}$
    \ENDIF
\ENDFOR

\STATE $P_{final} \leftarrow P_{initial} \cup P_{bridge}$
\RETURN $P_{final}$
\end{algorithmic}
\end{algorithm}

\section{Implementation Details}
\label{app:impl}

This section provides implementation details of TGS-RAG to ensure reproducibility, including hyperparameter settings, model choices, and system configuration. All parameters reported here correspond to the best-performing configuration identified through the sensitivity analysis described in Appendix~\ref{app:sensitivity_hparams}.

\subsection{Hyperparameters}

Table~\ref{tab:hyperparams} summarizes the hyperparameters used in all experiments unless otherwise specified. 
These parameters govern the bidirectional synergy between text and graph retrieval, including semantic--structural balancing, graph search behavior, and memory-based orphan entity bridging.

\begin{table}[h]
\centering
\caption{Hyperparameter Settings for TGS-RAG}
\label{tab:hyperparams}
\begin{tabular}{lcc}
\toprule
\textbf{Parameter} & \textbf{Symbol} & \textbf{Value} \\
\midrule
Chunk Synergy Weight & $\alpha$ & 0.5 \\
Text Confirmation Bonus & $\epsilon$ & 0.4 \\
Beam Search Width & $K$ & 20 \\
Search Depth & $d$ & 3 \\
Max Neighbors per Node & $N_{\max}$ & 30 \\
Top Orphan Entities Bridged & $k_o$ & 3 \\
Seed Density Bonus & $\gamma$ & 0.4 \\
Entity Degree Weight & $\lambda_e$ & 0.01 \\
Relation Degree Weight & $\lambda_r$ & 0.01 \\
Top Recommended Chunks & $k_r$ & 4 \\
Embedding Dimension & $Dim$ & 1024 \\
\bottomrule
\end{tabular}
\end{table}

The chunk synergy weight $\alpha$ controls the balance between semantic similarity and graph-based recommendation during text re-ranking, while the text confirmation bonus $\epsilon$ determines the strength of textual evidence in validating graph paths. 
The remaining parameters primarily affect search efficiency and structural bias and are shown to have limited sensitivity within reasonable ranges.

\subsection{Environment and Models}

\begin{itemize}
    \item \textbf{Large Language Model (LLM):} GPT-4o-mini is used for all LLM-invoked components, including entity extraction, knowledge graph construction, and final answer generation.
    \item \textbf{Embedding Model:} We employ the \textbf{Qwen3-Embedding-0.6B} model with an output dimension of 1024 for encoding queries, text chunks, and graph entities.
    \item \textbf{Database:} All text chunks, graph entities, relations, and their embeddings are stored in PostgreSQL with the \texttt{pgvector} extension, enabling unified vector-based retrieval over both structured and unstructured data.
    \item \textbf{Execution Environment:} Experiments are conducted on a standard Linux server with 32GB RAM. No GPU acceleration is required, as all model inference is performed via API-based services.
\end{itemize}

\subsection{Reproducibility Notes}

To ensure reproducibility and stable evaluation:
\begin{itemize}
    \item All entity extraction results from user queries are cached to eliminate stochastic variation from repeated LLM calls.
    \item Hyperparameter tuning is conducted on a dynamically balanced validation subset of MuSiQue, with deterministic sampling to guarantee consistent comparisons across configurations.
    \item All reported results use the same fixed hyperparameter configuration identified as optimal in the sensitivity analysis.
\end{itemize}

\section{Case Study: Overcoming Semantic Drift}
\label{sec:case_study}

To intuitively demonstrate the effectiveness of our bidirectional synergy, we analyze a representative multi-hop query from the MuSiQue dataset where baseline methods failed due to semantic ambiguity.

\begin{figure}[h]
    \centering
    \begin{tcolorbox}[breakable, enhanced, colback=gray!5, colframe=black!70, title=\textbf{Case Study: Semantic Drift vs. Structural Bridging}, fontupper=\scriptsize, boxsep=1mm, left=1mm, right=1mm, top=1mm, bottom=1mm, before skip=4pt, after skip=4pt]
    \small
    \textbf{Query:} \textit{"Which actress had roles in the movies 'Janie Jones' and 'Signs'?"}
    
    \rule{\linewidth}{0.4pt}
    
    \textbf{\textcolor{red}{Baseline (w/o Synergy):}}
    \begin{itemize}[leftmargin=*]
        \item \textbf{Retrieved Text:} Chunks related to \textit{"Caro Jones"}, \textit{"Shirley Jones"}, and \textit{"Madison Jones"}.
        \item \textbf{Failure Cause:} The vector retriever was misled by the high semantic density of the token "Jones," resulting in \textbf{Semantic Drift}. It retrieved entities sharing the surname but lacking logical connection to the movie "Signs".
        \item \textbf{Outcome:} "Cannot Answer" (Precision: 0.00).
    \end{itemize}
    
    \rule{\linewidth}{0.4pt}
    
    \textbf{\textcolor{blue}{TGS-RAG (Ours):}}
    \begin{itemize}[leftmargin=*]
        \item \textbf{Graph Path Discovery:} 
        \texttt{Janie Jones $\xrightarrow{cast}$ \textbf{Abigail Breslin} $\xleftarrow{cast}$ Signs}
        \item \textbf{Mechanism:} Although the text chunk for "Abigail Breslin" initially had a lower semantic score than "Caro Jones", the \textbf{Graph Channel} identified her as a \textbf{common neighbor} (bridge node) connecting both query entities.
        \item \textbf{Synergy:} The \textit{Graph-Guided Re-ranking} module boosted the priority of the "Abigail Breslin" chunk, suppressing the irrelevant "Jones" family chunks.
        \item \textbf{Outcome:} Correct Answer: "Abigail Breslin" (Recall: 1.00).
    \end{itemize}
    \end{tcolorbox}
    \caption{\textbf{Qualitative comparison on a multi-hop query.} While the baseline suffers from semantic drift (focusing on the surname "Jones"), TGS-RAG leverages the graph structure to identify the true bridge entity ("Abigail Breslin") that logically connects the two distinct movies, effectively correcting the retrieval focus.}
    \label{fig:case_study_22}
\end{figure}

As shown in Figure \ref{fig:case_study_22}, the baseline method falls into the "keyword trap," retrieving semantically similar but logically irrelevant entities (various people named "Jones"). In contrast, TGS-RAG successfully bridges the logical gap. The \textit{Semantic Beam Search} discovered that "Abigail Breslin" is the structural intersection of the two movies. Crucially, this structural signal allowed the system to "resurrect" the correct textual evidence that was otherwise buried by the vector retriever, validating the necessity of our graph-guided re-ranking mechanism.

\section{Hyperparameter Sensitivity Analysis}
\label{app:sensitivity_hparams}

We analyze the sensitivity of TGS-RAG to key hyperparameters that govern the bidirectional synergy between text and graph retrieval. 
All experiments are conducted on a dynamically balanced validation subset of MuSiQue, and performance is measured using Strict Hit Rate.

\paragraph{Effect of Chunk Synergy Weight $\alpha$.}
Varying the chunk synergy weight $\alpha$ exhibits a clear unimodal trend. 
Performance improves as $\alpha$ increases from 0.4 to 0.5 and degrades for larger values. 
This behavior reflects the trade-off between semantic similarity and graph-based structural guidance: smaller $\alpha$ underweights structural signals, while larger $\alpha$ overemphasizes graph recommendations at the expense of semantic relevance. 
The variance across runs is consistently small, indicating stable behavior around the optimal region.

\paragraph{Effect of Text Confirmation Bonus $\epsilon$.}
The text confirmation bonus $\epsilon$ shows a threshold-and-saturation pattern. 
Performance remains largely unchanged for small values of $\epsilon$ and peaks around $\epsilon = 0.4$, after which further increases yield no additional benefit. 
This confirms that textual evidence is most effective when used for validating graph paths rather than aggressively promoting them.

\paragraph{Beam Width and Search Depth.}
TGS-RAG demonstrates robustness to moderate variations in beam width $K$ and search depth $d$. 
Increasing the search depth from $d = 3$ to $d = 4$ yields no observable improvement in retrieval performance, suggesting diminishing returns from deeper exploration. 
This robustness indicates that the proposed memory-based orphan entity bridging mechanism effectively  recovers potentially valid reasoning chains discarded during search-time pruning.

\paragraph{Pairwise Sensitivity of $\alpha$ and $\epsilon$.}
We further analyze the joint effect of $\alpha$ and $\epsilon$ to assess parameter interaction stability. 
Results show that the optimal configuration ($\alpha = 0.5$, $\epsilon = 0.4$) is not an isolated point: neighboring combinations yield comparable performance with smooth degradation away from the optimum. 
This indicates that TGS-RAG does not rely on fragile parameter coupling and that its performance is stable across a contiguous region of the hyperparameter space.

Overall, the selected configuration ($\alpha = 0.5$, $\epsilon = 0.4$, $K = 20$, $d = 3$) achieves a favorable balance between effectiveness and efficiency and is used throughout all experiments.